# Deep Convolutional Framelet Denoising for Panoramic by Mixed Wavelet Integration

Masoud Shahraki Mohammadi, Seyed Javad Seyed Mahdavi Chabok

**Abstract**— Enhancing the quality and eliminating noise for the duration of preprocessing is a critical step in medical photo processing, in particular for X-ray photographs. These images are fashioned when photons collide with atoms, and variations in scattered noise absorption can lead to huge photo degradation. Such noise not most effective influences the clarity of the scientific images, however additionally frequently effects in repeat scans, thereby increasing the affected person's radiation exposure. Reducing picture noise has been a continual project on this subject. Traditional techniques like BM3D, low-skip filters, and Autoencoders had been hired for noise reduction with varying tiers of success. However, over the last decade, neural networks have proven super progress in surpassing those traditional techniques due to their sophisticated systems and higher accuracy.

One promising technique is the mixing of the Hankel matrix with neural networks. This matrix is designed to split neighborhood and non-nearby components in the photograph, allowing greater specific noise discount. This separation is normally completed in the use of a non-nearby matrix created with strategies like wavelets or DCT. In this paper, we recommend a progressive method that combines the Daubechies (D4) wavelet with an u-Net neural network architecture. T4 wavelet is selected for its advanced power awareness residences, and it is carried out at every level of the u-Net version to decorate the noise reduction procedure.

The effectiveness of this approach become evaluated in the use of PSNR and SSIM metrics, and the effects have been proven across various waveforms. Our experiments show that the integration of this wavelet improves noise discount overall performance by 0.5% to 1.2% as compared to other models, demonstrating the capacity of a one-wave network for greater correct medical picture denoising.

## I. Introduction

Classification and low-level vision computers like segmentation [1], super-resolution [2], [3], and denoising [4] have both benefited from deep learning techniques. The theoretical foundations of its conquest have been extensively explored [5], [6], and it operates by mimicking the brain's approach of using multiple layers of neurons to learn high-level features.

Deep learning has seen success in many areas over the past ten years. Kang et al. [7] pioneered the systematic proposal of applying deep convolutional neural networks (CNNs) to computed tomography (CT) scans acquired at low doses. They demonstrated that the use of a CNN with directional wavelets significantly improves the denoising process. The streaking artifacts caused by sparse projection views present globalized disturbances that pose a considerable challenge for removal when using conventional denoising CNNs, despite originating from low-dose artifacts due to reduced tube currents [8], [9]. Han et al. [10] and Jin et al. [11] proposed the adoption of residual learning with U-Net [1] to effectively mitigate the global streaking artifacts caused by sparse projection views. Wang et al. [12] were the first to use compressed sensing MRI with deep learning in MRI (CS-MRI). They used images from a downscale reconstruction to train a deep neural network to learn a fully sampled reconstruction.

The biggest problem for signal processing is the integration of deep learning and signal processing theories. Wavelets [13] have been extensively studied for numerous image processing applications as a practical signal representation theory, leveraging the effects of energy compaction in wavelet bases.

For numerous image processing applications, techniques such as non-local means [14] and BM3D [15] have demonstrated superior performance. These non-local image processing methods have proven to be exceptionally effective. However, sometimes

---

[1] Masoud Shahraki Mohammadi is with the Department of the computer engineering, Islamic Azad University of Mashhad, Mashhad, Iran (e-mail: m.shahraki@mshdiau.ac.ir)

Seyed Javad Seyed Mahdavi Chabok is with the Department of the computer engineering, Islamic Azad University of Mashhad, Mashhad, Iran (e-mail: mahdavi@mshdiau.ac.ir)

blindly, deep learning outperforms traditional signal processing that is based on mathematics.

Hankel matrices are employed in encoders and decoders by Yin et al. [16], who pioneered the application of classical signal processing theory, grounded in mathematics, to the field of deep learning.

We diverge from these prevalent viewpoints in this paper and extend the Hankel matrix of a deep network to a deeper network. Utilizing Daubechies wavelets and taking into consideration their constraints in pinpointing local features, researchers attempted to reduce the level of noise in panoramic images.

This research is focused on minimizing the system response time to approach as near to zero as possible, addressing a critical obstacle prevalent in medical processes. Hence, the restriction of neural network expansion will be considered a default.

## II. Basic concepts

Successful scientific research requires an understanding of the diverse initial paths that the research may follow. Recognizing these paths enables researchers to adapt flexibly to findings and challenges that arise during the course of their study.

So we present some fundamental ideas in this section.

### A. Hankel matrix decomposition and framelet convolution

One of the characteristics of the Hankel matrix is its matrix decomposition. The terms "local" and "non-local" refer to this decomposition in the context of framelet convolution [3].

It is assumed that the Hankel matrix $H_d(f)$ with amplitude $r < d$ has a single value decomposition (1) for the input signal for the input signal $f \in R^n$:

$$H_d(f) = U \Sigma V^T \tag{1}$$

Where $U = [u_1 \ldots u_r] \in R^{n \times r}$ and $V = [v_1 \ldots v_r] \in R^{d \times r}$ are the base vectors of the single left and right values, respectively, and $\Sigma \in R^{n \times r}$ form the diameter of the matrix, which contains single values. Multiplying $U^T$ and $V$ gives the (2):

$$\Sigma = U^T H_d(f) V \tag{2}$$

The Hankel matrix's left and right matrices can be thought of in terms of the following equation:

$$C_{ij} = \phi_i^T H_d(f) \psi_j = \langle f . \phi_i \circledast \psi_j \rangle . \; i = 1. \ldots . n \; . j = 1. \ldots 1 \, d \tag{3}$$

Which $C_{ij}$ represents the interaction of f with the non-local BIOS $\phi_i$ and the local BIOS $\psi_j$. We arrived at the convolution framelet expansion phase of the signal expansion.

$$f = \frac{1}{d} \sum_{i=1}^{n} \sum_{j=1}^{d} \langle f . \phi_i \circledast \psi_j \rangle \phi_i \circledast \psi_j \tag{4}$$

Equation (4) For each $f \in R^n$ can expand with convolution frame $\phi_i \circledast \psi_j$ and $\langle f . \phi_i \circledast \psi_j \rangle$ conduct signal expansion. Although the local and non-local framelet coefficients in (1) and (2) are not as distinct from one another as they are in (2), Yin suggests that training $\Psi$ as $\Phi$ can be used to obtain the local data. A critical step in the process is choosing a non-local BIOS. Selecting a non-local BIOS is an important factor in the process.

### B. Selecting non-local BIOS

With more energy compression, single value decomposition (SVD) performs the left and right suitable vector well [17][18] [19]. It has complicated calculations, however. It has complicated calculations, however. This complexity often requires significant computational resources, especially for large datasets. Nevertheless, the benefits of SVD in data reduction and noise removal frequently outweigh these computational challenges.

### C. Haar

The conversion of the Haar wavelet yields the BIOS of Haar, which is defined as follows.

$$\Phi = [\Phi_{low} \; \Phi_{high}] \tag{5}$$

Where the high-pass and low-pass operators $\Phi_{low} \; \Phi_{high} \in R^{n \times \frac{n}{2}}$ are defined as follows:

$$\Phi_{low} = \frac{1}{\sqrt{2}}[1\ 0\ 1\ 0\ 0\ 0\ 1\ 1\ \cdots\ 0\ 0\ 0\ \vdots\ \ddots\ \vdots\ 0\ 0\ 0\ 0\ \cdots\ 1\ 1\ ]$$
$$\Phi_{high} = \frac{1}{\sqrt{2}}[1\ 0\ -1\ 0\ 0\ 0\ 1\ -1\ \cdots\ 0\ 0\ 0\ \vdots\ \ddots\ \vdots\ 0\ 0\ 0\ 0\ \cdots\ 1\ -1\ ] \quad (6)$$

Each column comprises two non-zero elements, indicating that a Haar wavelet expansion is not expected, as illustrated in Figure 1. As a result, various rotations and transitions are implemented on the wavelet, adding a 90-degree extra step for the user.

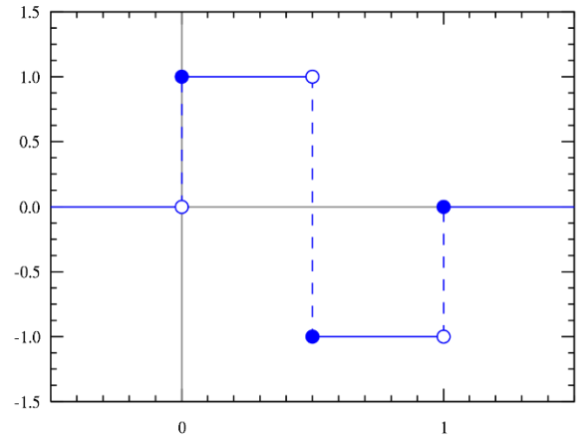

Fig. 1. Haar wavelet

### D. DCT

The Discrete Cosine Transform (DCT) represents a finite sequence of data points in terms of a sum of cosine functions oscillating at varying frequencies. These transformations are widely used in science and engineering. They are essential for the lossy compression of audio data, as with MP3s, and images, such as JPEGs, where small, high-frequency components can be eliminated. Additionally, Discrete Cosine Transforms (DCTs) are employed in spectral methods for the numerical resolution of partial differential equations.

Given that a smaller number of cosine functions is needed to approximate a standard signal compared to sine functions, employing the cosine function is essential for efficient compression. Cosmic parts also have more definite boundary conditions than differential functions.

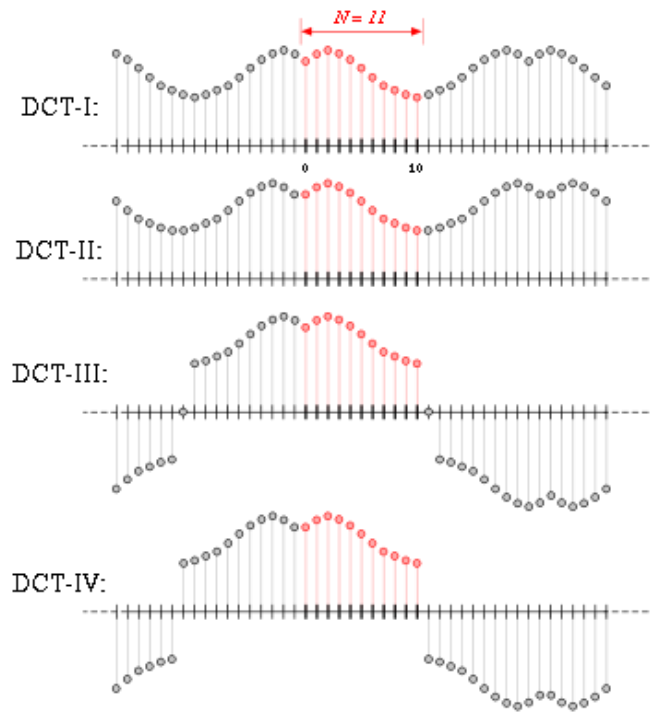

Fig. 2. Types of DCT

## III. U-Net deep neural network

The u-net deep learning network is based on convolutional networks [1], [21]. This network comprises two parts: pooling (left) and unspooling (right). The grid consists of two layers of 3 * 3 convolution, each completed with Relu and Max pooling 2 × 2 with step 2 for sampling. In each layer, the maximum accumulation of the number of feature channels will be doubled. This operation is conducted in reverse using the same parameters on the network; however, the sole distinction is that convolutional aggregation is employed instead of a comprehensive collection. A total of 23 layers have been used in this network.

In this section, instead of the cumulative layers, the Haar wavelet will be used as the master wavelet. We will investigate this transformation by integrating various types of complete wavelets with the outcomes derived from the wavelet energy. The integration aims to enhance the feature extraction process by capitalizing on the spatial-frequency characteristics inherent in the Haar wavelet. This approach promises to refine the network's accuracy in recognizing patterns and textures within the input data.

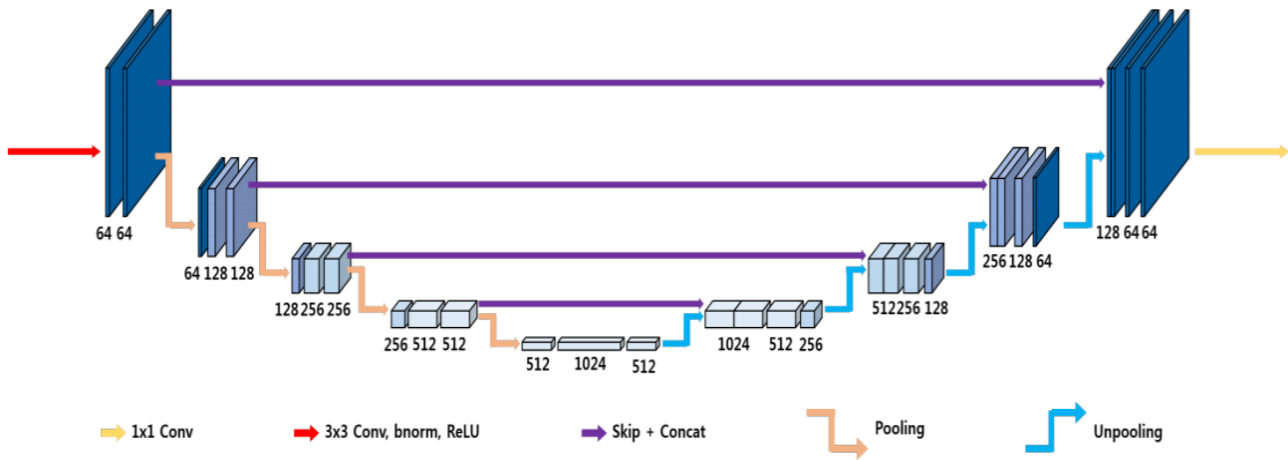

Fig. 3. Deep neural network [11] U-net structure, which is used for our comparative studies.

### A. Haar wavelet

In [20], high accuracy has been achieved by using the Haar wavelet. However, as stated at the outset of the study, enhancing the noise reduction rate remains a critical challenge in the preprocessing phase. To tackle this challenge, innovative combinations of wavelet transforms with advanced filtering techniques have been proposed to further improve signal clarity. These methods aim to minimize the presence of unwanted noise while preserving the essential details necessary for accurate analysis.

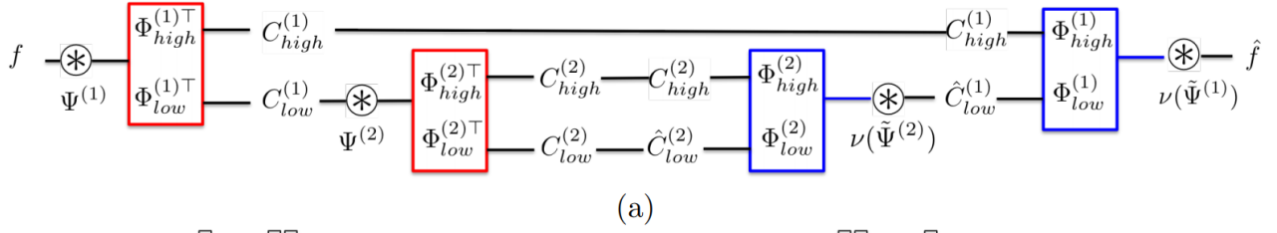

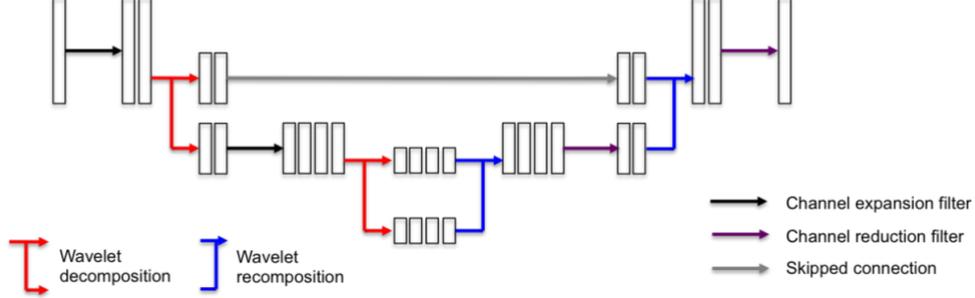

Fig. 4. Transformation of haar wavelet in deep neural network [20]

Fig. 4 shows an overview of haar wavelet placement in a U-net-based neural network. Each of the boxes is as follows:

$$C_{high}^{(1)} = \Phi_{high}^{(1)^T} H_{d_{(1)}}(f)\Psi^{(1)} = \Phi_{high}^{(1)^T}\left(f \circledast \underline{\Psi}^{(2)}\right)$$
$$C_{low}^{(1)} = \Phi_{low}^{(1)^T} H_{d_{(1)}}(f)\Psi^{(1)} = \Phi_{low}^{(1)^T}\left(f \circledast \underline{\Psi}^{(1)}\right)$$
(7)

And also in the second layer:
$$C_{low}^{(2)} = \Phi_{low}^{(2)^T} H_{d_{(2)}|p_{(2)}}\left(C_{low}^{(1)}\right)\Psi^{(2)} = \Phi_{low}^{(2)^T}\left(C_{low}^{(1)} \circledast \underline{\Psi}^{(2)}\right)$$
$$C_{high}^{(2)} = \Phi_{high}^{(2)^T} H_{d_{(2)}|p_{(2)}}\left(C_{high}^{(1)}\right)\Psi^{(2)} = \Phi_{high}^{(2)^T}\left(C_{high}^{(1)} \circledast \underline{\Psi}^{(2)}\right)$$
(8)

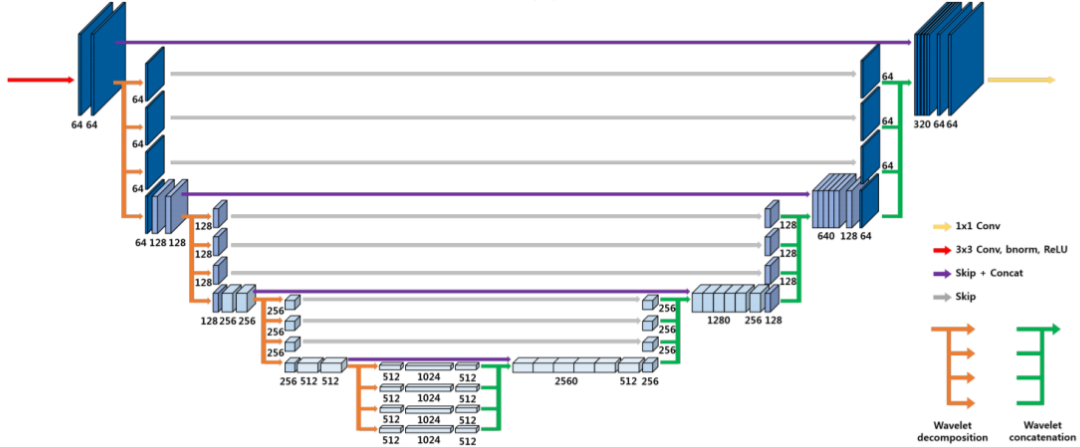

Fig. 5. Haar wavelet in u-net [1]. Show the outline of the haar wavelet in each layer

## IV. Proposed Method

This section explores the proposed method for accessing a deeper network by combining the Haar and Db2 wavelets. The hybrid approach aims to capitalize on Haar's simplicity and Db2's capability to capture more complex features in the data. Experiments show that this combination greatly enhances how well the model works, especially when it's important to keep detailed information and clear signals.

According to [22], the sampling rate of Daubechies wavelets surpasses that of Haar wavelets, leading to enhanced accuracy achieved through the combination of Db2 and Haar wavelets. Specifically, merging these two types of wavelets enables a more adaptable representation of a signal, thus allowing for a more precise and efficient process of pattern recognition. The experimental results confirm that the hybrid model improves the overall performance and significantly reduces noise sensitivity, which is essential for applications involving real-world data where noise is inherently present.
However, due to the u-net and wavelet network settings, this combination was not possible, which is as follows:

1. step size = 2; Because the haar wavelet is two-dimensional (Fig. 6)
2. Padding = zero; Adding any data to the input will cause noise
3. Network input depth = 64

The first reason is the ability to compare with older networks, and the second is that, according to (7), each decomposition step doubles the network depth and halves the size of the input image. Given that there are four levels of decomposition and the input image cannot exceed 1024 by 1024 pixels due to GPU memory constraints.     These settings make Db2, which consists of four elements, with a step size of two at the end of the two pixels image less than the Db2 function.

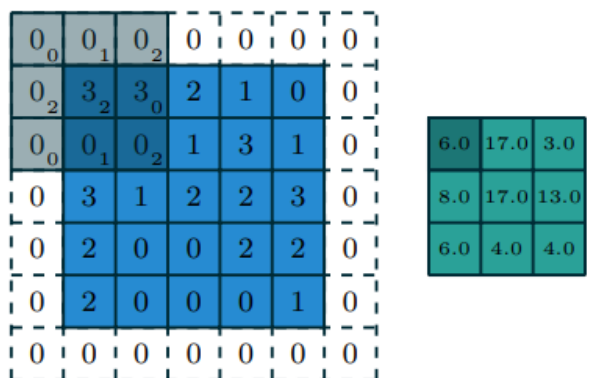

Fig. 6. How to step size works at neural network output [23]

## V. Integrated neural network settings

According to the settings of the previous section and the limitations mentioned, the proposed solution is as follows:
The whole neural network is divided into decomposition and concatenation, which are symmetrical to each other. There are four steps for placing wavelets, each with its wavelet and settings. In this way, there will be no fixed step size in the whole network, and in each step, the step size is selected according to the same wavelet. This operation has limitations such as network deepening and loss of image information in the first layers if we use a more complex wavelet.

Therefore, not every arbitrary combination of the network can be used.
1. The step size in the Haar wavelet will be two and db2 will be four.
2. Padding = zero
3. Network input depth = 64, but since the use of other wavelets deepens the network, more depth can be used.
4. Adam-based network training
5. Training rate 0.0001
6. Every 25 stages of the training rate is divided into two

For example, you can see a combination of Haar wavelet and Db2. Given that the wavelet consists of two steps, we systematically assign the number two to each deployed layer. Consequently, within this context, the wavelet Db2 is referred to as "four." In Fig. 7, it will be denoted as "4422." The first and second layers are Db2 with step four, and the following two layers are Haar with step two. Consequently, the designation "4422" signifies a hierarchical wavelet transformation. In this structure, the first two layers supply a greater level of detail or frequency information due to the employment of Db2 on a more refined scale, which incorporates step four. This is succeeded by a broader or averaged transformation, implemented through Haar wavelets at a coarser scale, utilizing step two. This combination enables a multi-resolution analysis that captures both fine and coarse features within the signal or image being processed.
Changing the step size in each layer depends on the type of wavelets in the same filter layer. In this way, any step depending on the desired layer can be selected. This flexibility in choosing the step size allows for an optimization of the signal processing according to the specific requirements of the task at hand. By adjusting these parameters, the system can be finely tuned to enhance the detection of subtle nuances or to focus on broader trends within the data.

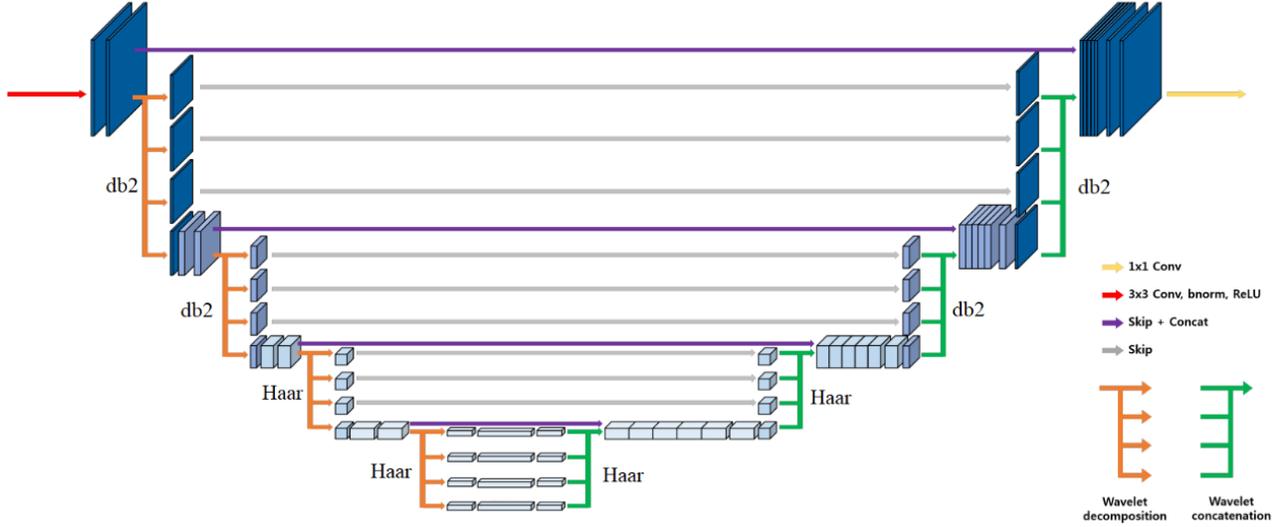

Fig. 7. wavelet integration Haar and db2

# VI. Experimental result

This section explores the outcomes of image denoising. In particular, we will focus on our deep convolutional framelets using Db2 wavelets. By leveraging the sparse representation of Daubechies 2 (Db2) wavelets, we significantly enhance the denoising capabilities of framelets, ultimately achieving superior image reconstruction quality. Experimental results demonstrate the efficacy of our approach in preserving edge details while removing noise.

### A. Comparison of wavelet integration and work done

In this section, we will compare the performance of U-Net, Haar wavelet, and various combinations of Haar wavelet and Db2. The comparison criteria are PSNR (9) and SSIM (10) [24][25], [26].

$$\text{PSNR} = 20\ log_{10}\left(\frac{MAX_Y}{\sqrt{MSE(\hat{X}).Y}}\right) \qquad (9)$$

$$\text{MSE} = \frac{1}{mn}\sum_{i=0}^{m-1}\sum_{j=0}^{n-1}\left(f(i.j) - H(i.j)\right)^2 \qquad (10)$$

Where MAX is the maximum amount of pixels in the image. j, i show the rows and columns of the image, f and H are the original high-resolution and reconstructed images, respectively. The simplest, widely used evaluation criterion is the mean square error. This criterion is attractive because it is computationally simple, has a clear physical meaning, and is mathematically appropriate for optimization. The higher the PSNR, the better the image reconstruction.

$$\text{SSIM} = \frac{(2\mu_{\hat{X}}\mu_Y + c_1)(2\sigma_{\widehat{XY}} + c_2)}{(\mu_{\hat{X}}^2 + \mu_Y^2 + c_1)(\sigma_{\hat{X}}^2 + \sigma_Y^2 + c_2)} \qquad (11)$$

Where $\mu_{\hat{X}} = \frac{1}{N}\sum_{i=1}^{N} I_i$ is the average intensity in input I and $\sigma_{\widehat{XY}} = \frac{1}{(N-1)}\sum_{i=1}^{N}\left(I_i - \mu_{\hat{X}}\right)\left(I_i - \mu_{\hat{i}}\right)$ is the standard deviation. $c_1$ and $c_2$ are two constants whose result will be output stability. When it is higher, it will be more efficient.

### B. Datasets

The primary dataset, comprising over 500,000 panoramic X-rays, was created by Dr.Hoseini Zarch, an expert in Oral and Maxillofacial Radiology, utilizing Computed Radiography technology. For comparison with the previous method, we use three standard datasets listed.

Set12 dataset: This dataset consists of 12 black and white images with sizes of 256 by 256 and five 512 by 512 images.

Set14 dataset: This dataset consists of 14 color images of different sizes.

BSD68 database: This database consists of 68 black and white photos with different sizes.

## C. Results obtained

The results were obtained on a system with the specifications of i5-7400 3.00GHz and GTX 1080Ti 11 Gb.
The unit of σ is decibels, and the noise entering the system is almost the same as the noise coming from panoramic graphics, which has been obtained experimentally. The amount of noise added to the photos σ = 30.

Noise model Speckle $\quad\quad\quad I(i.j) = S(i.j) * N(i.j) \quad\quad\quad\quad\quad\quad\quad\quad\quad\quad\quad\quad\quad\quad\quad\quad$ (12)

Where S is the input signal and N is the product of the mean multiplication in standard deviation $\sigma$. [27]

Table I
Performance comparison in terms of PSNR

| Dataset | Input | U-net | Haar | 4422 |
|---|---|---|---|---|
| Set12 | 18.7805 | 28.7188 | 29.5126 | 29.8963 |
| Set14 | 18.8264 | 28.4994 | 28.5978 | 28.9345 |
| BSD68 | 18.8082 | 27.0842 | 27.8836 | 27.8536 |

Table II
Performance comparison in terms of SSIM

| Dataset | Input | U-net | Haar | 4422 |
|---|---|---|---|---|
| Set12 | 0.2942 | 0.8194 | 0.8280 | 0.8283 |
| Set14 | 0.3299 | 0.7966 | 0.7866 | 0.8001 |
| BSD68 | 0.3267 | 0.7787 | 0.7761 | 0.7795 |

As can be seen, using the haar wavelet at the beginning of the network causes sampling and convolution to be done more accurately in the early stages and then continues with the db2 wavelet. You will see other combinations below.

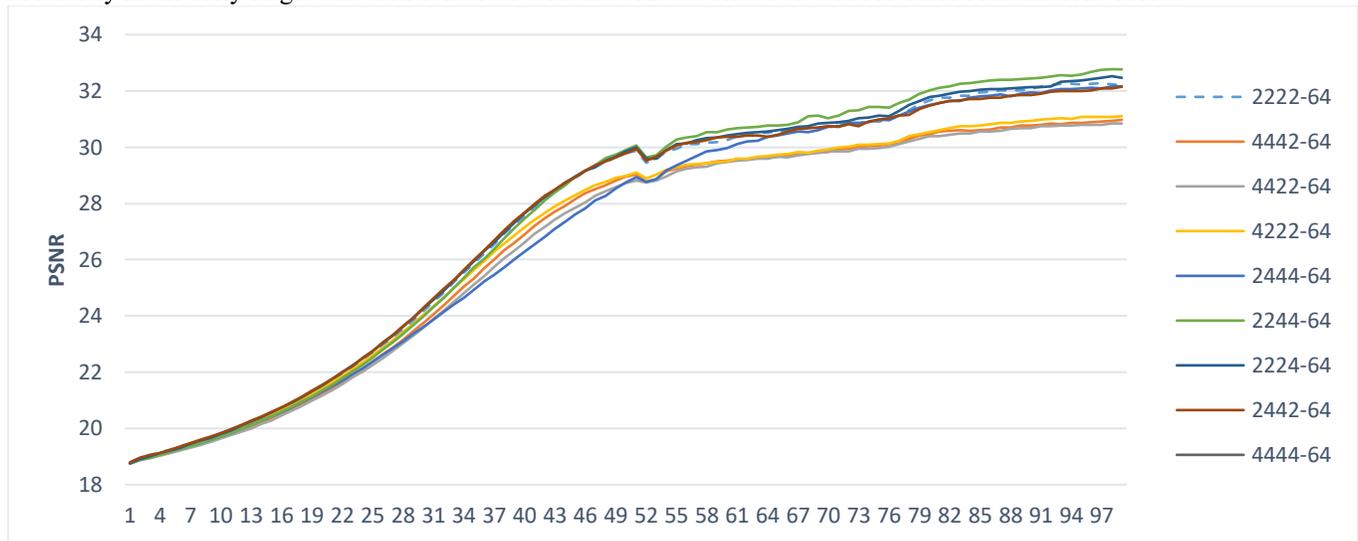

Fig. 8. Comparison between output PSNR corresponding to different model in dataset Set12

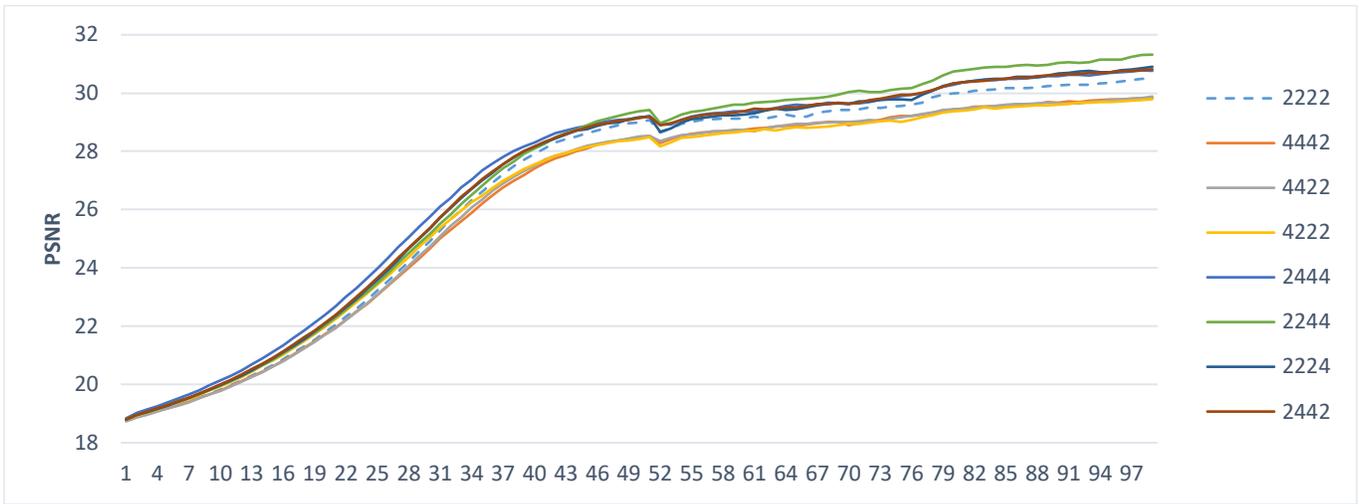
Fig. 9. Comparison between output PSNR corresponding to different model in dataset Set14

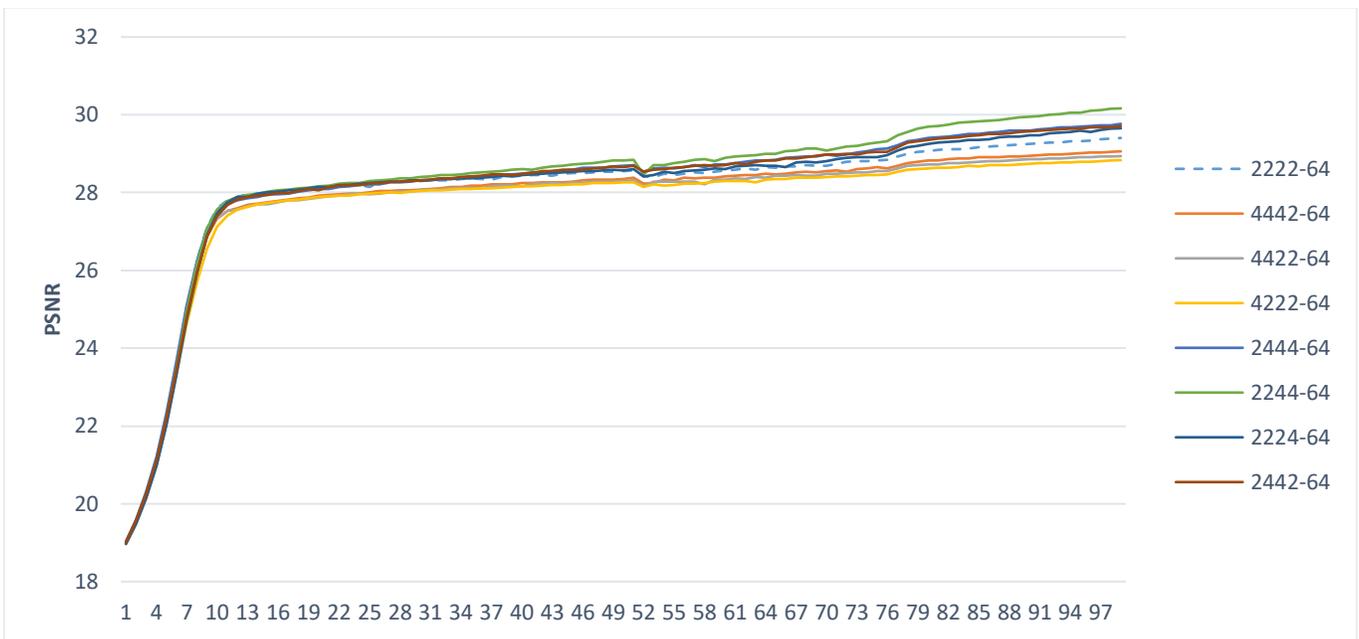
Fig. 10. Comparison between output PSNR corresponding to different model in dataset BSD68

The common point of all outputs is on stages 50 and 51 and later, which is justified by the drop. The reason is to apply optimizer changes every 25 steps, which is more extensive in the second application than before. These changes cause the network to deviate more than the previous limits, and this drop is observed in the output.

In Set12, the convergence is swifter owing to its black-and-white composition and standard dimensions, which eliminate the need for padding that introduces additional noise. Set 14 follows closely behind in this regard. Consequently, this suggests that grayscale or monochrome images offer less complexity for the model to process, leading to more rapid improvements and stability during training. Therefore, the data preprocessing stage becomes crucial, especially when dealing with colored and high-resolution images, to ensure uniformity and better performance.

In BSD68, more weights are updated due to the more significant number of images per cycle of the neural network. This higher number of network scans also means faster network divergence.

In all datasets, layout Fig. 7 or 4422 has exhibited the best performance. One reason is that initially, through a longer step (Db2) process, a more comprehensive set of information about the image is processed. Subsequently, utilizing a shorter step (db1) allows for the finer details of the photo to be processed with greater precision. This dual-step process ensures that not only is the broad structure of the image captured accurately, but also the minute details are finely delineated, enhancing the overall quality of the image processing. Therefore, the integration of Db2 and db1 steps proves to be a highly efficient approach for optimizing network scans in image processing tasks.

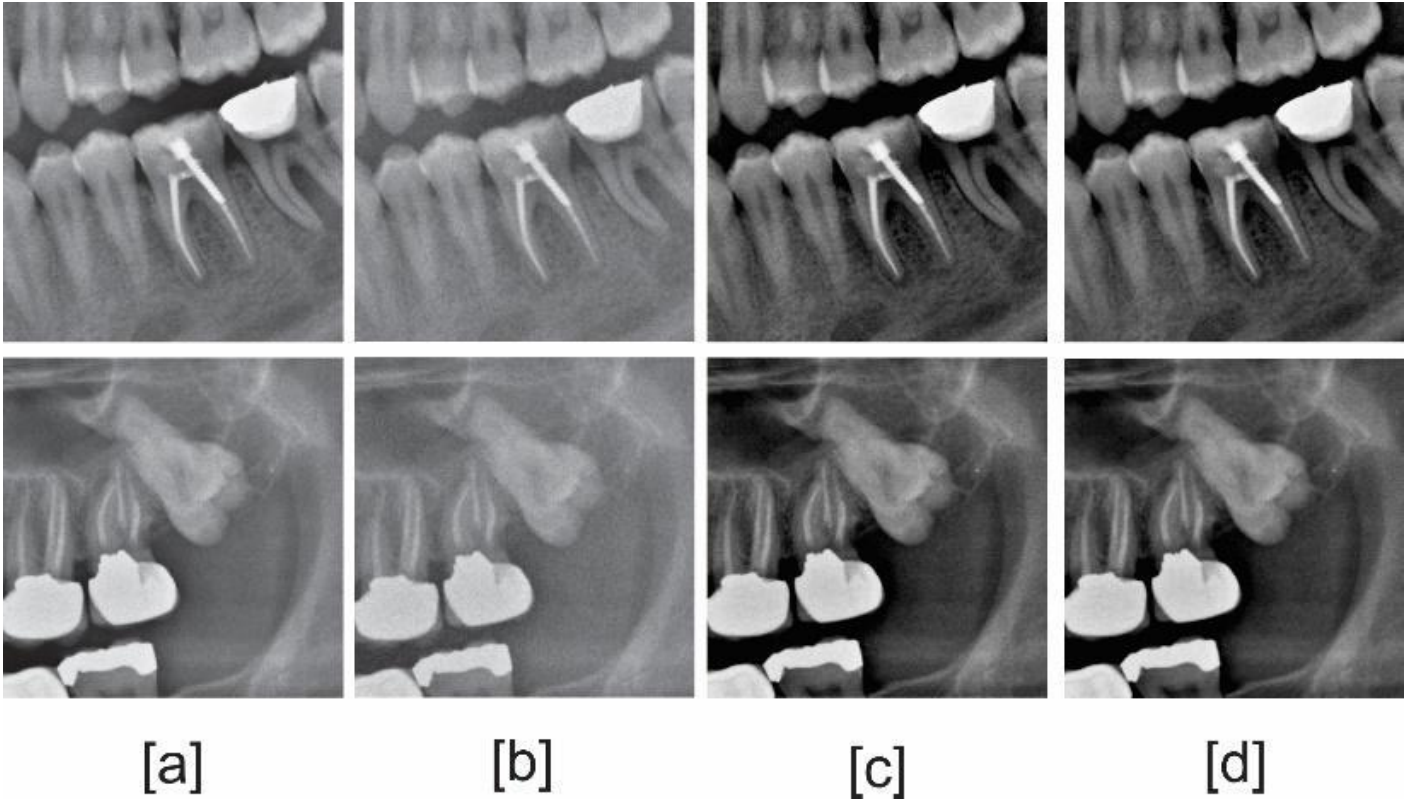

Fig. 11. Denoising results of two images from Dataset Master. [a] Original. [b] Noisy with 30 $\sigma$. [c] Haar wavelet (2222). [d] Mixed wavelet (4422)

## VII. Conclusion

This method was chosen to enhance denoising on account of neural networks' demonstrated superiority over prior techniques, as indicated by the results. Neural networks have more power due to the high repetition of the algorithm on the database. One of these networks is the autoencoder network, which decomposes into four layers and merges into four. The architecture effectively captures the underlying patterns and structures within noisy datasets and reconstructs the signal with impressive fidelity. Autoencoders are particularly effective at reducing dimensionality and aiding in denoising by learning to ignore noise while retaining valuable signals.

The Henkel matrix is defined in such a way that it facilitates its use as a convolutional operation within a neural network. One of its notable properties is the singular value analysis, which facilitates the matrix's approximation. This property requires two orthogonal matrices called local and non-local matrices. Obtaining these two matrices is the subject of this research. One way is to get a non-local matrix of wavelets. By benefit the energy of wavelets and integrating them, we have introduced an innovative method for noise reduction through neural networks, utilizing a framelet approach. We first showed that the Henkel matrix could be divided into local and non-local. Framelets can act like the classic encoder-decoder layer in the neural network. By disseminating this concept and introducing a novel wavelet to the network layers, we outperformed the results obtained with the Haar wavelet. This improvement is evident in both qualitative and quantitative measurements in various signal and image processing applications. The proposed method improves noise reduction while keeping important details and signal sharpness, essential for accurate interpretation and analysis.